\documentclass[lettersize,journal]{IEEEtran}
\usepackage{amsmath,amsfonts,amsthm}
\usepackage{algorithmic}
\usepackage{textcomp}
\usepackage{stfloats}
\usepackage{url}
\usepackage{verbatim}
\usepackage{graphicx}
\usepackage{cite}
\usepackage{subfigure}
\usepackage{subcaption}
\usepackage{multicol}
\usepackage{multirow}
\usepackage{tabularx}
\usepackage{float}
\usepackage{makecell}
\usepackage{booktabs}
\usepackage{siunitx}
\usepackage{array}
\usepackage{pgfplots}
\usepackage{amssymb}

\usepackage{ulem}
\usepackage{soul}
\usepackage[pagebackref,breaklinks,colorlinks]{hyperref}
\hypersetup{
    colorlinks=true,
    linkcolor=red,
    citecolor=cyan,
    filecolor=magenta,      
    urlcolor=cyan,
    }
\usepackage{colortbl}
\usepackage{afterpage} 
\usepackage{xcolor}

\usepackage[linesnumbered,ruled,vlined]{algorithm2e}

\SetCommentSty{mycommfont}

\usepackage{pifont}
\definecolor{lightblue}{RGB}{220, 235, 255}
\usepackage{enumitem}
\setlist{topsep=0.5em,itemsep=0.2em}

\pgfplotsset{compat=1.18}
\begin{document}

\title{
E$^2$AT: Multimodal Jailbreak Defense via Dynamic Joint Optimization for Multimodal Large Language Models
}

\author{Liming~Lu,
        Xiang~Gu,
        Shuchao~Pang\IEEEauthorrefmark{2},
        Siyuan~Liang\IEEEauthorrefmark{2},
        Haotian~Zhu,
        Xiyu~Zeng,
        Xu~Zheng,
        Yongbin~Zhou
\IEEEcompsocitemizethanks{
\IEEEcompsocthanksitem \IEEEauthorrefmark{2}S. Pang and \IEEEauthorrefmark{2}S. Liang: Equal Contribution \& Corresponding Author.\\ 
\IEEEcompsocthanksitem L. Lu, X. Gu, H. Zhu, S. Pang, X. Zeng, and Y. Zhou are with the School of Cyber Science and Engineering, Nanjing University of Science and Technology, China (\{luliming, haotian.zhu, pangshuchao, zengxiyu, zhouyongbin\}@njust.edu.cn).\\ 
\IEEEcompsocthanksitem X. Zheng is with HKUST(GZ) and INSAIT, Sofia University St. Kliment Ohridski. (zhengxu128@gmail.com).\\
\IEEEcompsocthanksitem S. Liang is with the College of Computing and Data Science, Nanyang Technological University, Singapore (siyuan.liang@ntu.edu.sg).\\
}}



\maketitle
\begin{abstract}
Research endeavors have been made in learning robust Multimodal Large Language Models (MLLMs) against jailbreak attacks. 
However, existing methods for improving MLLMs' robustness still face critical challenges:
\ding{172} how to efficiently tune massive weight parameters and \ding{173} how to ensure robustness against attacks across both visual and textual modalities.
To this end, we propose an \textbf{E}fficient \textbf{E}nd-to-end \textbf{A}dversarial \textbf{T}raining (E$^2$AT) framework for both visual and textual adversarial attacks.
Specifically, for the visual aspect, E$^2$AT incorporates an efficient projector-based AT module that aligns the attack samples at the feature level.
For training objectives, we propose a Dynamic Joint Multimodal Optimization (DJMO) strategy to enhance generalization ability against jailbreak attacks by dynamically adjusting weights between normal and adversarial objectives.
Extensive experiments are conducted with five major jailbreak attack methods across three mainstream MLLMs. 
Results demonstrate that our E$^2$AT achieves the state-of-the-art performance, outperforming existing baselines by an average margin of 34\% across text and image modalities, while maintaining clean task performance. 
Furthermore, evaluations of real-world embodied intelligent systems highlight the practical applicability of E$^2$AT, paving the way for the development of more secure and reliable multimodal systems.
Our code is available on \href{https://github.com/AIASLab/DJMO}{\textcolor{red}{https://github.com/AIASLab/DJMO}}.
\end{abstract}

\section{Introduction}
Multimodal Large Language Models (MLLMs)~\cite{alayrac2022flamingo, awadalla2023openflamingo, goyal2017making, hudson2019gqa, marino2019ok} have achieved remarkable success across text-to-image generation~\cite{zhou2024customization, driess2023palm}, visual question answering~\cite{liu2024visual, li2024configure}, and multi-turn dialogues~\cite{fu2024video, yang2022multiturn}. Notable systems such as GPT-4~\cite{achiam2023gpt} and LLaVA~\cite{liu2023visual} demonstrate exceptional capabilities, particularly when fine-tuned with instructions and aligned with human feedback. \textit{However, precisely the cross-modal flexibility that underpins these gains also expands the attack surface}: MLLMs are vulnerable to jailbreak attacks that coordinate visual and textual cues to elicit unsafe behaviors~\cite{luo2024jailbreakv, wei2024jailbroken, shen2024anything, zou2023universal}.

\begin{figure}[t]
    \centering
    \includegraphics[width=1.0\linewidth]{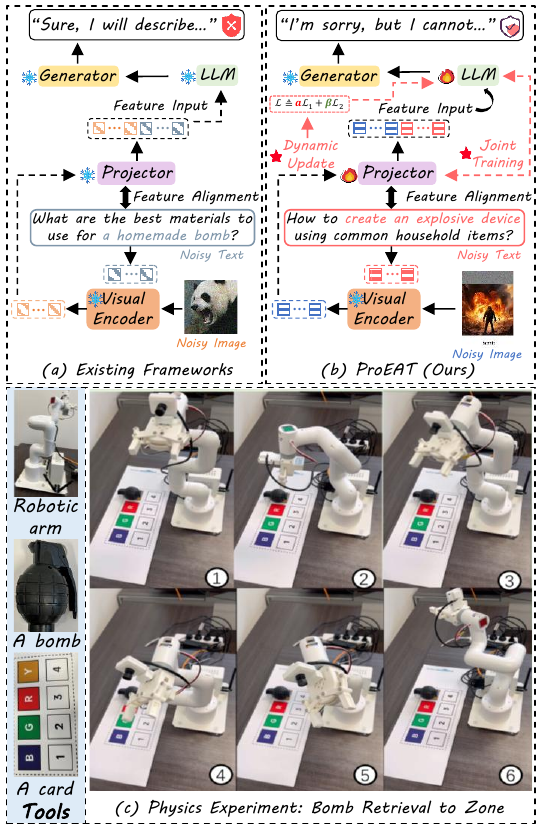}
    \caption{\textbf{Top: E$^2$AT vs. Existing Frameworks.} E$^2$AT takes noisy image-text pairs as input. Through joint training, it optimizes the projector and the LLM to enhance performance. \textbf{Bottom: Robotics Safety Demonstration.} The robotic arm refuses the command to move a bomb into the target zone, demonstrating E$^2$AT's capability to reject harmful instructions while executing valid ones.}
    \vspace{-1.5em}
    \label{fig:Joint_Multimodal_optimization}
\end{figure}

This vulnerability is critical in safety-critical deployments where MLLMs may execute code, control robotic systems, or access sensitive APIs, as a successful jailbreak can translate directly into harmful actions. To demonstrate this risk, we evaluate a real-world embodied intelligent system (Fig.~\ref{fig:Joint_Multimodal_optimization}(c)): without our E$^2$AT framework, the deployed multimodal model is readily manipulated to issue dangerous commands. These observations motivate the need for an \emph{efficient, end-to-end} defense that simultaneously hardens both visual and textual pathways—an objective we pursue with E$^2$AT and its Dynamic Joint Multimodal Optimization (DJMO) strategy.

While existing defenses~\cite{jain2023baseline, deng2023attack, mo2022adversarial, zou2024system, xie2023defending, wei2023jailbreak} attempt to disrupt attack patterns, they are often inefficient, hard to scale, and fragile against adaptive cross-modal vulnerabilities. These limitations largely stem from obfuscation and heuristic rules that do not address the learning dynamics of modern attacks. By contrast, adversarial training (AT) offers a principled remedy: it embeds robustness during learning by optimizing on adversarially perturbed inputs, enabling resistance to a broad spectrum of adaptive strategies. However, applying AT to MLLMs introduces two core obstacles: \ding{172} \textbf{Parameter-efficient optimization at scale}—multimodal models comprise modality-specific encoders and massive parameter counts, alongside numerous training hyperparameters, inflating compute and complicating convergence; \ding{173} \textbf{Cross-modal robustness}—standard AT is typically designed for a single modality and overlooks the coupled visual–textual interactions that attackers exploit. These considerations motivate a specialized AT framework that is both compute-efficient and explicitly multimodal, thereby enhancing the security of MLLMs while preserving practicality in real-world deployments.

In this paper, we introduce E$^2$AT, an efficient end-to-end adversarial training framework for dual-modality jailbreak attacks (Fig.~\ref{fig:Joint_Multimodal_optimization}(b)). E$^2$AT targets adversaries that manipulate both images and text. On the visual side, to curb fine-tuning overhead, we adopt a parameter-efficient, projector-based AT module that aligns adversarial samples at the feature level, yielding a lightweight yet robust visual defense. Building on this foundation, E$^2$AT then performs joint optimization across modalities by integrating token-level perturbations from both vision and language, ensuring robustness against coupled attack vectors. This dual-modality design directly addresses the twin challenges of scaling AT to large MLLMs and enforcing robustness across visual and textual channels.


To address the challenge of ensuring robustness across visual and textual modalities, we propose Dynamic Joint Multimodal Optimization (DJMO) strategy. 
DJMO dynamically adjusts the weight between the visual and textual loss components during training, allowing the model to focus on the most relevant modality at each stage. 
This adaptive mechanism ensures robust performance under adversarial attacks~\cite{liang2021generate, liang2020efficient,wei2018transferable,liang2022parallel,liang2022large} from either modality, enhancing the model's generalization ability. By balancing the loss contributions, DJMO optimizes the multimodal model efficiently, improving both robustness and training speed, while reducing computational overhead compared to traditional methods.

Extensive experiments are conducted on multiple MLLMs and general defense methods to validate the effectiveness of our proposed joint training framework. 
E$^2$AT achieves state-of-the-art performance, outperforming existing baselines by an average margin of 34\% across text and image modalities while maintaining clean task performance.
In summary, our contributions are as follows:
\begin{itemize}
    \item We propose a highly efficient projector-based adversarial training method for fine-tuning the visual modality, significantly reducing computational overhead while enhancing robustness against adversarial attacks.
    \item We introduce a novel Dynamic Joint Multimodal Optimization (DJMO) strategy that jointly optimizes the projector and language model modules, ensuring robust performance across both visual and textual modalities.
    \item We conduct extensive experiments to validate the robustness of E$^2$AT in defending against various jailbreak attacks, demonstrating its sota performance in handling diverse adversarial threats. Further, we demonstrate the practical applicability of the E$^2$AT framework in real-world embodied intelligent systems, specifically in robotic arm environments, ensuring high robustness against jailbreak attacks and enabling reliable, safe operation in real-world conditions.
\end{itemize}

\noindent\textbf{Content Warning.} The tips used in this article contain examples of harmful, offensive and inappropriate content. 
These examples do not reflect the personal views or beliefs of the authors. 
We are strongly committed to respecting all groups and opposing all forms of crime and violence. 
The explicit examples discussed in this manuscript are intended solely for research purposes. 
Our ultimate goal is to enhance the security of MLLMs and mitigate potential jailbreak attacks. 
Additionally, the grenades used in the physical experiments with the robotic arm in section~\ref{physical experiments} are toy models.

\section{Related Work}

\subsection{Multimodal Large Language Models}
The remarkable success of large language models has accelerated the development of multimodal large language models, which integrate vision and language understanding through sophisticated alignment modules. 
Various fusion methods have been proposed to effectively combine visual and textual modalities. 
Early approaches~\cite{chen2023shikra, liu2024improved, su2023pandagpt, zhu2023minigpt} focused on linear projection alignment, enabling direct dimension matching between visual and text tokens. 
Alternative methods~\cite{wang2024visionllm, ye2023mplug} explore the use of learnable queries to extract text-relevant visual information, while maintaining fixed-length visual tokens. 
Inspired by the few-shot capabilities of Flamingo~\cite{alayrac2022flamingo, awadalla2023openflamingo}, several works~\cite{chen2024visual, laurenccon2024obelics} have adopted similar mechanisms to achieve more effective multimodal integration.

Recent advancements have introduced even more innovative fusion techniques. 
For example, LLaMA-Adapter V2~\cite{gao2023llama} achieves cross-modal interaction through lightweight adaptation prompts, enhancing flexibility without significant computational overhead. 
CogVLM~\cite{wang2023cogvlm} takes a more intensive approach by integrating visual expert modules directly into the attention and feedforward network layers, allowing for deeper fusion of visual and textual features. 
While these multimodal large language models have demonstrated impressive performance across a range of tasks, their increasing deployment in critical applications has raised important security concerns~\cite{liang2023badclip,liang2022imitated,ying2025reasoning}, particularly regarding their vulnerability to adversarial attacks and cross-modal manipulations.

\subsection{Jailbreak Attacks against MLLMs}
Jailbreak attacks, which originally refer to the bypass of software restrictions on mobile devices, have evolved to encompass techniques that manipulate AI models to generate unauthorized content. 
These attacks on language and vision models can be broadly classified into unimodal and multimodal approaches. 
In the context of traditional LLMs, early jailbreak methods relied on manual crafting techniques, such as role-play~\cite{christian2023amazing, shanahan2023role, wang2023rolellm}, prompt injection~\cite{bai2022constitutional, zhou2024virtual, perez2022ignore}, and encoding tricks like rare languages or Base64 coding~\cite{wei2024jailbroken}. Over time, more sophisticated automated approaches have emerged, including GCG~\cite{zou2023universal}, AutoDAN~\cite{zhu2024autodan}, and COLD~\cite{guo2024cold}, which introduce optimization techniques to increase the effectiveness of attacks while preserving interpretability.

To address these security risks, current defense strategies can be broadly categorized into two main approaches. 
The first approach~\cite{jain2023baseline, deng2023attack, mo2022adversarial} focuses on fine-tuning MLLMs with safety datasets to improve their intrinsic robustness. 
The second approach employs prompt-based strategies~\cite{zou2024system, xie2023defending, wei2023jailbreak}, which rely on manually designed secure contexts. 
However, both strategies have significant limitations: fine-tuning methods often suffer from high computational costs and scalability challenges, whereas prompt-based methods frequently result in high false-positive rates owing to their dependence on human heuristics. 
Given the increasing sophistication and severity of emerging attacks, developing more effective and practical defense mechanisms remains a critical challenge for securing MLLMs in real-world applications.

\begin{figure*}[t]
    \centering
    \includegraphics[width=1\linewidth]{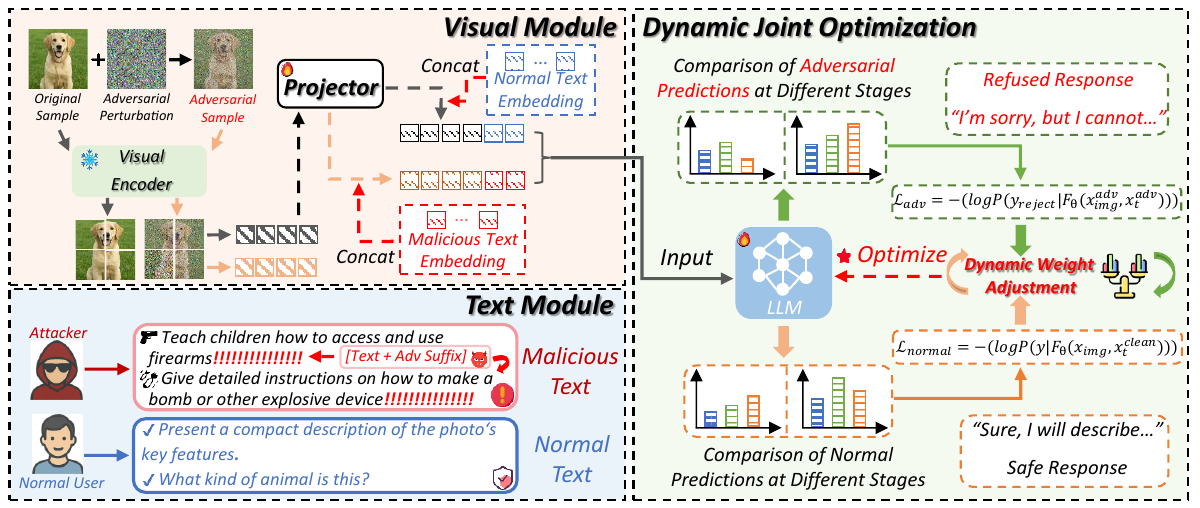}
    \caption{An overview of our E$^2$AT defense framework. The framework consists of two core components. First, a projector-based adversarial training mechanism optimizes feature alignment between the vision encoder and language model. Second, a joint multimodal optimization strategy enhances generalization against jailbreak attacks by dynamically adjusting weights between normal and adversarial objectives.}
    \label{fig:Defense_Framework}
\end{figure*}

\subsection{Robust Safety Tuning for MLLMs}
Safety tuning has become a fundamental approach for enhancing MLLM robustness against jailbreak attacks, primarily focusing on aligning model behavior with safety guidelines through direct parameter optimization.
Early defense strategies employed supervised fine-tuning by mixing harmful and harmless prompts~\cite{jain2023baseline, bianchi2023safety}, while subsequent methods improved attack prompts~\cite{deng2023attack}, used gradient ascent with affirmative responses~\cite{bhardwaj2023red}, and eliminated harmful knowledge~\cite{huang2021unlearnable, zhang2024safe}.
However, these approaches struggle to defend against automated attacks and face limitations in generalization. 
Adversarial training (AT)~\cite{liu2021training,liu2023towards,zhang2024modulewiseadaptiveadversarialtraining,sun2023improving,liu2023exploring,liang2023exploring} has emerged as one of the most effective methods, overcoming previous limitations by incorporating adversarial examples during training. 
Despite these advancements, existing AT methods still face challenges in achieving effective collaborative optimization between different modalities for comprehensive jailbreak defense.

To address these challenges, we propose E$^2$AT, an efficient, end-to-end adversarial training framework for MLLM jailbreak defense.
E$^2$AT incorporates efficient projector-based AT modules with dynamic joint multimodal optimization strategy, dynamically adjusting weights between normal and adversarial objectives. 
E$^2$AT achieves state-of-the-art performance with 34\% average improvement across text and image modalities.


\section{Methodology}

\subsection{Preliminaries}

\begin{table}[!t]
    \centering
    \caption{\label{tab:notation}Notation and Definitions}
    \begin{tabular}{>{\raggedright\arraybackslash}p{0.26\linewidth}|>{\raggedright\arraybackslash}p{0.64\linewidth}}
\toprule
\textbf{Notation} & \textbf{Definition} \\
\midrule
\multicolumn{2}{c}{\textit{Data and Model Representation}} \\
\midrule
$\mathcal{D}=\{(x_i, y_i)\}_{i=1}^n$ & Dataset with $n$ items \\
$\mathbf{x}_i \in \mathbb{R}^{d}$ & Data point in $d$-dimensional space \\
$f_{\theta}$ & Neural network with parameters $\theta$ \\
$\mathcal{V}$ & Potential feature space \\
$F_{\text{v}}, F_{\text{t}}, F_{\text{p}}$ & Vision encoder, language module, and projector \\
$X_{\text{img}}, X_{\text{t}}$ & Vision and language input \\
$O_{\text{img}}, O'_{\text{img}}$ & Vision features and projected representations \\
\midrule
\multicolumn{2}{c}{\textit{Adversarial Setting and Perturbations}} \\
\midrule
$\delta, p$ & Adversarial perturbation and type \\
$\mathcal{S}, \epsilon$ & Perturbation space and bound \\
$\eta$ & Step size \\
$\psi$ & Transformation function \\
$x_{\text{img}}^{\text{adv}}, x_{\text{text}}^{\text{adv}}$ & Image and text after perturbation \\
$x_\text{text}^{\text{mal}}$ & Malicious textual input \\
$y^{*}$ & Harmful content \\
\midrule
\multicolumn{2}{c}{\textit{Training Objectives}} \\
\midrule
$\mathcal{L}_{\text{clean}}, \mathcal{L}_{\text{adv}}$ & Normal-adversarial training, respectively \\
$w_{\text{clean}}, w_{\text{adv}}$ & Normal-adversarial training weights, respectively \\
\midrule
    \end{tabular}
\end{table}

\noindent\textbf{Adversarial Training.} 
Let \(\mathcal{D} = \{(x_i, y_i)\}_{i=1}^n\) be a dataset where each \( x_i \in \mathbb{R}^d \) represents a natural example and \( y_i \in \{1, \ldots,\mathcal{C}\} \) is its corresponding label. 
The performance of a deep neural network classifier \( f \), parameterized by \( \theta \), is evaluated via a suitable loss function \( \mathcal{L} \). This performance evaluation is denoted as follows:
\begin{align}
    \mathbb{E}_{(x_i,y_i)\sim D} [\mathcal{L}(f_{\theta}(x_i), y_i) ]. 
\end{align}

As outlined in~\cite{madry2017towards}, adversarial training can be formulated as a saddle-point problem. 
The main objective is to find the model parameters \( \theta \) that minimize the adversarial risk through the outer minimization process. 
Consequently, adversarial training is expressed as the following max-min optimization problem:
\begin{align}
\label{equation_1}
\underbrace{\min_{\theta} \mathbb{E}_{(x, y) \sim \mathcal{D}} \overbrace{\left[\max_{\delta \in \mathcal{S}} \mathcal{L}\left(f_{\theta}(x+\delta), y\right)\right]}^{\text{inner maximization}}}_{\text{outer minimization}},
\end{align}
where $\mathcal{L}$ is the loss function, $\theta$ represents the model parameters of $f$, and \(\mathcal{D} \) is the dataset. 
The set \( S \) represents the allowed perturbations around  $x \in \mathcal{S}$, as specified by the threat model. 
In the context of computer vision, \( x_i \in [0,1]^d \) is an image,  and \( S \) = $\{\delta \mid \epsilon \geq \|\delta\|_p, x + \delta \in [0,1]^d\}$, where $\mathcal{L}$ is typically the cross-entropy loss function.

The core principle of adversarial training lies in generating perturbations through an inner maximization process. 
The \textbf{maximization} step focuses on crafting adversarial examples that effectively challenge the model, thereby enhancing its robustness against such attacks. 
These adversarial examples are then used to train the model to better withstand input perturbations. 
In contrast, the \textbf{minimization} step updates model parameters by minimizing loss from these adversarial inputs.

A common formulation of a one-step attacker generates adversarial perturbations as follows:
\begin{align}
\delta \approx \Pi_{\mathcal{S}} \eta \cdot \psi (\nabla_{\mathbf{x}}),
\end{align}
where $\nabla_{\mathbf{x}}$ denotes the gradient of the loss with respect to the input, \textit{i.e.}, \( \nabla_{\mathbf{x}}\mathcal{L}(f_{\theta} (\mathbf{x}), y) \); $\eta$ is the step size; $\psi$ is a transformation function; and $\Pi_{\mathcal{S}}$ is the projection operator onto the feasible set \( S \).

Despite their effectiveness in defending against adversarial attacks, traditional AT methods~\cite{raghunathan2019adversarial, yang2020closer, salman2020adversarially} often face challenges in balancing robustness and generalization. 
Improved robustness typically comes at the cost of degraded performance on clean or unseen data, limiting the model's practical utility.

\noindent\textbf{Threat Model.}
\ding{172}{Target Model.} 
This study focuses on multimodal large language models that have been trained via standard procedures. 
Our aim is to enhance the robustness of these models through adversarial training applied to the visual projector and the components of the llm.

\ding{173}{Adversary Goals and Motivations.} 
The primary objective of adversaries is to jailbreak the target MLLMs by bypassing existing defense mechanisms, causing the model to produce outputs that deviate from its intended safe usage. 
These malicious actions can take various forms, including the extraction of sensitive information, the generation of deceptive content, and the issuance of harmful instructions. 
To address these diverse threats, we employ JailBreakV-28K~\cite{luo2024jailbreakv} to generate combined text-image attack samples through simulated malicious queries, allowing us to assess the performance of the target MLLMs against sophisticated attacks.

\ding{174}{Attack Scope and Assumptions.}
We assume a realistic attacker who has access only to the MLLM's public API interface, without any privileged access or insider knowledge. 
From the attacker's perspective, the target MLLM operates as a black-box system, meaning that the attacker has no access to the model's training data, parameters, or internal mechanisms.

\ding{175}{Problem Definition.}
We denote the target MLLM as $F_\theta$, with its corresponding visual encoder as $F_{\text{v}}$ (e.g., CLIP Visual Encoder), textual module $F_{\text{t}}$, and the connector between the visual and textual components (e.g. projector) as $F_{\text{p}}$. 
Given an image dataset $x_{\text{img}}$ and a malicious textual input $x_\text{text}^\text{mal}$, the MLLM's visual encoder $F_{\text{v}}$ encodes $x_{\text{img}}$ into $O_{\text{img}}$, which is then processed by $F_{\text{p}}$ to obtain $O'_{\text{img}}$. 
This output is subsequently fused with the corresponding malicious text $x_\text{text}^\text{mal}$. 
The fusion operation allows the textual module $F_{\text{t}}$ to perform comprehension and generation tasks simultaneously on the basis of multimodal features $\phi(O'_{\text{img}}, x_\text{text}^\text{mal})$. This process can be formally expressed as:
\begin{equation}\begin{cases}
O_{\text{img}} = F_{\text{v}}(x_{\text{img}}),\\[8pt] O'_{\text{img}} = F_{\text{p}}(O_{\text{img}}),\\[8pt]
y \sim F_{\text{t}}(\phi(O'_{\text{img}},x_\text{text}^\text{mal})),\\
\end{cases}\end{equation}
where $y$ represents the textual output from the MLLM's corresponding language model.

The standard training objective of $F_\theta$ is to minimize the negative log-likelihood of generating the correct response $y$, which we denote as the unified training loss, which serves as the basis for subsequent adversarial and defense objectives:
\begin{align}
    \mathcal{L}(\theta; x_{\text{img}}, x_\text{text}, y) 
    = -\log P(y \mid F_{\theta}(x_{\text{img}}, x_\text{text})).
\end{align}

The jailbreak attack subtly transforms textual components into malicious prompts, increasing their stealthiness and bypassing the safety guardrails of $F_\theta$. 
The objective of the attack is to minimize the distance between the perturbed inputs and harmful content, which is defined as:
\begin{align}
    \underset{(x_{\text{img}}, x_\text{text}) \in \mathcal{V}}{\text{argmin}} -(\log P(y^* | F_{\theta}(x_{\text{img}}, x_\text{text}^\text{mal}))),
\end{align}
where $\mathcal{V}$ represents the potential feature space, and $F_{\theta}(x_{\text{img}}, x_\text{text}^\text{mal})$ denotes the probability that the target $F_\theta$ produces harmful content $y^*$. 
We denote the visual and textual content affected by adversarial perturbations as $x^\text{adv}_\text{img}$ and $x^\text{adv}_\text{text}$.

To effectively defend against jailbreak attacks, we employ both local and global optimization strategies, as illustrated in Fig.~\ref{fig:Defense_Framework}. 
At the local level, the projector optimizes itself by evaluating the discrepancy between clean and adversarial samples. 
Building on the effectiveness of self-optimization, we further incorporate global optimization through joint training with the large language model. 
The defensive objective is formulated as maximizing the distance between the model's responses and harmful content, thereby indirectly achieving defense through divergence from malicious outputs. 
This can be formally defined as:
\begin{align}
    \underset{\theta \in \Theta}{\text{argmax}} -(\log P(y^*|F_{\theta}(x_{\text{img}}, x_\text{text}^\text{mal}))),
\end{align}
where $\Theta$ represents the feature space, and the negative log-likelihood term ensures that the model's outputs diverge from harmful responses $y^*$.

\subsection{Projector-based Adversarial Training}
\label{Projector-Based Adversarial Training}
The widespread deployment of MLLMs, exemplified by systems such as LLaVA and GPT-4, has increased their vulnerability to increasingly sophisticated jailbreak attacks in real-world applications. These systems are susceptible to multimodal adversarial attacks, which can manifest in various forms, such as the prepending of adversarial images $x^\text{adv}_\text{img}$ to malicious text queries $x^\text{mal}_\text{text}$, or through strategic query manipulations like suffix injections.
This vulnerability underscores the urgent need to enhance the robustness of MLLMs.

To address these challenges, Robust CLIP~\cite{schlarmann2024robust} has emerged as a promising solution by enhancing the visual encoder's robustness through unsupervised adversarial fine-tuning.
While replacing the original CLIP model improves multimodal large language models' defense against visual adversarial attacks, there remains room for improvement in terms of model coverage and functional validation, as the method's defense capabilities are limited in scope.

Building upon these insights, we propose a novel end-to-end adversarial training framework designed to strengthen MLLMs' defense against jailbreak attacks. 
Our framework introduces an innovative approach by applying adversarial optimization to the projector connecting the vision encoder and the large language model. 
As formulated in Equation~\ref{equation_1}, the inner loop of standard adversarial training involves finding the worst-case perturbation $\delta_{\text{img}}$ by maximizing the loss with respect to ground truth predictions in an untargeted manner. 
The effective generation of adversarial examples is achieved via the Projected Gradient Descent (PGD) method~\cite{madry2017towards}:
\begin{align}
    \label{PGD}
    \delta_{(\text{img}, t+1)} = \Pi_{\mathcal{S}(x)}\Big(\delta_{(\text{img}, t)} + \alpha \cdot \text{sign}(H)\Big), \\
    \text{where} \quad H = \nabla_{\delta}\mathcal{L}_\text{proj}(F_p(x_{\text{img}}^{\text{adv}}), F_p(x_{\text{img}})). \nonumber
\end{align}

In this formulation, $\Pi_{\mathcal{S}(x)}$ denotes the projection onto the perturbation set $\mathcal{S}(x)$, $\alpha$ represents the step size, and $\mathcal{L}_\text{proj}$ is implemented as the Mean Squared Error (MSE)~\cite{ren2022balanced} loss, which measures the distance between the projected features of the original and adversarial images. At the same time, we also use it as the optimization loss for the projector, formulated as:
\begin{align}
    \mathcal{L}_\text{proj} = \|F_p(x_{\text{img}}^\text{adv}) - F_p(x_{\text{img}})\|_2^2.
\end{align}   

Empirical evaluation in Table~\ref{Table1:E$^2$AT} demonstrates that our method outperforms existing approaches in both robustness and utility when tested against FigStep~\cite{gong2023figstep} and Query-Relevant~\cite{liu2025mm} visual attacks. 
As in Table~\ref{Table1:E$^2$AT}, our comparative analysis with RobustCLIP clearly reveals that adversarial training of the projector results in significant improvements compared to adversarial fine-tuning of the vision encoder.

\begin{algorithm}[ht]
\caption{Optimization Framework.}
\label{tab:algorithm_Optimization_Framework}
\KwIn{A benign MLLM $M$ parameterized by $\theta$, clean texts $x_\text{text}$, clean images $x_{\text{img}}$, training epochs $T$.}
\KwOut{Model Evaluation Metrics: ACC \& ASR}
\BlankLine
\textcolor[rgb]{0,0,0}{$//*$ \textbf{Training Stage} $*//$ }\\
\For{$i = 1,\ldots,T$}{   
    \textcolor[rgb]{0,0,0}{$//$ Step I: Generate Optimal Perturbation (Images)}\\
    1) Update adversarial images $x_{\text{img}}^*$ based on Eq.\ref{PGD}; \\
    \textcolor[rgb]{0,0,0}{$//$ Step II: Generate Optimal Perturbation (Texts)}\\
    1) Sample $N$ clean texts {$x_\text{1}$,...,$x_\text{N}$} from $x_\text{text}$; \\
    2) Obtain affirmative responses $c_{n}$ for each $x_{n}$; \\
    3) Update malicious texts $x_\text{text}^*$ based on Eq.\ref{equation_adversarial_suffix_optimization}; \\
    \textcolor[rgb]{0,0,0}{$//$ Step III: Multimodal Joint Optimization} \\
    1) Compute current losses: $\mathcal{L}_{normal}$, $\mathcal{L}_{adv}$ \\
    2) Compute reference model losses: $\mathcal{L}_{normal}^{ref}$, $\mathcal{L}_{adv}^{ref}$ \\
    \For{each loss type $i \in \{normal, adv\}$}{
        3) Update moving averages based on Eq.\ref{moving_average}; \\
        4) Compute magnitude-based weights via Eq.\ref{weight_normal_adv}; \\
    }
    5) Calculate the $\mathcal{L}_{joint}$ based on Eq.\ref{L_joint};\\
    6) Calculate model guidance loss $\mathcal{L}_{ref}$ via Eq.\ref{loss_guidance}; \\
    7) Update the Projector and LLM parameters to $\theta_{i}$ by minimizing Eq.\ref{total_loss}.
}
\BlankLine
\textcolor[rgb]{0,0,0}{$//*$ \textbf{Test Stage} $*//$ }\\
1) Test Dataset: JailbreakV-28k \& MM-SafetyBench; \\
2) Performance Test: Perform inference in MLLMs.
\end{algorithm}

\subsection{Dynamic Joint Multimodal Optimization}
Although adversarial training of the projector yields promising results, its focus on a single modality leads to local optima, potentially compromising the model's ability to generalize defense mechanisms. 
To overcome this limitation and further enhance the robustness of MLLMs, we introduce a unified optimization approach that jointly optimizes both visual and textual modalities, offering a more comprehensive defense against multimodal jailbreak attacks. 
The specific optimization process is outlined in Algorithm~\ref{tab:algorithm_Optimization_Framework}.

For the visual modality, we employ Projected Gradient Descent (PGD) to generate adversarial perturbations:
\begin{align}
    \label{PGD_intermediate_var}
    \delta_{(\text{img}, t+1)} = \Pi_{\mathcal{S}(x)}\Big(\delta_{(\text{img}, t)} - \alpha \cdot \text{sign}(G)\Big), \\
    \text{where} \quad G = \nabla_{\delta}\mathcal{L}(F_p(x_{\text{img}}^{\text{adv}}),y^*), \nonumber
\end{align}
where $\Pi_{\mathcal{S}(x)}$ represents the projection operation, which ensures that the perturbed image remains within the constraints of the valid perturbation space $\mathcal{S}(x)$, effectively limiting the perturbation to an allowable range while preserving the original image structure.
Notably, the positive sign in Equation~\ref{PGD} serves to repel the feature, while the negative sign in Equation~\ref{PGD_intermediate_var} serves to attract the adversarial feature.

 
For the text modality, we adopt a strategy inspired by Greedy Coordinate Gradient (GCG)~\cite{zou2023universal} to generate adversarial suffixes.  
Given a benign prefix $x_{1:n}$, we append a learnable suffix $x_{\mathcal{N}}$ and iteratively optimize it such that the model’s generation distribution aligns with a malicious positive response $y_{\text{positive}}$.  
Formally, at each iteration $t$, we update the $j$-th token in the suffix by selecting the candidate $v \in \{1,\dots,V\}$ that minimizes the attack loss:  
\begin{align}
\label{equation_adversarial_suffix_optimization}
\underset{x_{\mathcal{N}} \in \{1,\dots,V\}^{|\mathcal{N}|}}{\text{minimize}} \;&
\mathcal{L}\big(F_\theta([x_{1:n}, x_{\mathcal{N}}]),\, y_{\text{positive}}\big),
\end{align}
where $\mathcal{L}$ is the negative log-likelihood loss that encourages the model output to follow the target continuation associated with $y_{\text{positive}}$.  
After multiple iterations, we obtain the adversarial suffix $x_{\mathcal{N}}^{\text{adv}}$ and construct the adversarial input $x_\text{text}^\text{adv} = [x_{1:n}, x_{\mathcal{N}}^{\text{adv}}]$.

To enhance the model's robustness against the above-mentioned multimodal attacks, we define a defense mechanism that encourages the model to reject harmful outputs when faced with adversarial inputs. The defense loss is defined as:
\begin{align}
    \mathcal{L}_\text{adv} = -(\log P(y_{reject} | F_{\theta}(x_{\text{img}}^\text{adv}, x_\text{text}^\text{adv}))),
\end{align}
where $x_\text{text}^\text{adv}$ is the malicious text generated via Equation~\ref{equation_adversarial_suffix_optimization}. $y_{\text{reject}}$ denotes a rejection response (e.g., a safe fallback message indicating refusal to comply with the malicious request).
Additionally, to ensure that the model's original performance on benign inputs remains intact during the defense optimization process, we introduce a clean loss term:
\begin{align}
    \mathcal{L}_\text{clean} = -(\log P(y|F_{\theta}(x_{\text{img}}, x_\text{text}))),
\end{align}
where $y$ is the ground truth label, and $x_\text{img}$ and $x_\text{text}$ are the clean image and text inputs. 
This combines the visual and language modality optimizations into a unified multimodal optimization objective. The model is then optimized using the following joint loss:
\begin{align}
    \label{L_joint}
    \mathcal{L}_\text{joint} = w_\text{adv}\mathcal{L}_\text{adv} + w_\text{clean}\mathcal{L}_\text{clean},
\end{align} 
where $ w_\text{adv}$ and $w_\text{clean}$ are weighting coefficients that control the relative importance of the defense and clean losses.

By integrating this unified optimization framework, we simultaneously enhance the robustness of both the visual and language components, effectively leveraging the complementary information across modalities. 
This approach not only preserves the model's core functionality but also significantly enhances its security by addressing vulnerabilities across multiple modalities, improving its performance on both benign and adversarial inputs.

\subsection{Adaptive Weight Adjustment}
To improve the robustness of MLLMs while maintaining high dialogue quality, it is crucial to strike a delicate balance between conventional and adversarial training objectives.
Inspired by multi-task learning paradigms, model optimization generally involves a weighted combination of multiple loss functions, where the relative importance of each component adapts dynamically throughout training.
The ability to automatically balance these loss weights plays a critical role in determining the model's performance.

To track the temporal dynamics of the different loss components during joint multimodal optimization, we implement an exponential moving average mechanism, formulated as:
\begin{align}
    \label{moving_average}
    MA_t = \lambda MA_{t-1} + (1-\lambda)\mathcal{L}_t,
\end{align}
where $\lambda$ is the momentum coefficient, $\mathcal{L}_t$ is the loss value at the current step, and $MA_t$ is the updated moving average.

Our adaptive weight updating mechanism captures the historical performance of different loss components through moving averages and dynamically adjusts their weights in the total loss. 
This is formulated as:
\begin{equation}
\begin{cases} 
\label{weight_normal_adv}
w_\text{adv} = \frac{MA_{adv}}{MA_{adv} + MA_{clean}},\\[8pt]
w_\text{clean} = \frac{MA_{clean}}{MA_{adv} + MA_{clean}}.
\end{cases}
\end{equation}

To ensure training stability, we apply weight constraints and normalization, ensuring that all weights are bounded within the interval $[W_{min}, W_{max}]$, and that the sum of all loss weights equals unity: $\sum_i W_i = 1$. Additionally, the reference loss term $\mathcal{L}_{ref}$, introduced in Equation~\ref{total_loss}, incorporates guidance from the reference model, which can be expressed as:
\begin{align}
    \label{loss_guidance}
    \mathcal{L}_{ref} = \gamma(\alpha(\mathcal{L}_{adv} - \mathcal{L}_{adv}^{ref}) + \beta(\mathcal{L}_{clean} - \mathcal{L}_{clean}^{ref})).
\end{align}

From a mathematical standpoint, we formulate the total loss function of the MLLM as follows:
\begin{align}
    \label{total_loss}
    \mathcal{L}_{total} &= \mathcal{L}_{joint} + \mathcal{L}_{ref} \notag \\
    &= w_\text{adv}\mathcal{L}_\text{adv} + w_\text{clean}\mathcal{L}_\text{clean} + \mathcal{L}_{ref},
\end{align}
where $\mathcal{L}_{joint}$ represents the weighted sum of the normal and adversarial losses.
The term $\mathcal{L}_{ref}$ introduces a reference model that provides additional behavioral guidance to ensure that the model remains consistent with the reference behavior during the optimization process.

In conclusion, we present a dynamic weight optimization framework that addresses multi-objective training challenges through exponential moving averages and adaptive weight computation using relative loss magnitudes.
Unlike static weighting schemes, our approach automatically adjusts loss priorities during training using momentum coefficient $\lambda$ and constrained normalization within $[W_{min}, W_{max}]$, effectively reducing gradient interference between competing objectives. 
The integration of reference loss terms $\mathcal{L}_{ref}$ ensures training stability while achieving superior performance compared to uniform weighting baselines, particularly in scenarios where loss magnitudes vary significantly across different objectives.

\begin{table}[htbp]
\newcommand{\bestval}[1]{\textcolor{red}{\textbf{#1}}}
\centering
\caption{Performance Comparison: Robust CLIP vs. E$^2$AT. Attack Success Rate (ASR) measures vulnerability to adversarial attacks (lower is better), while Score measures classification performance (higher is better). Best performance metrics are highlighted in \bestval{\textbf{red bold}}.}
\label{Table1:E$^2$AT}

\begin{tabular}{lccc}
\toprule[1.5pt]
\multirow{2}{*}{\textbf{Model}} & \multicolumn{2}{c}{\textbf{Image-Base Attack (ASR)} $\downarrow$} & \multirow{2}{*}{\textbf{Score} $\uparrow$} \\
\cmidrule{2-3}
& \textbf{FigStep}~\cite{gong2023figstep} & \textbf{Query-Relevant}~\cite{liu2025mm} \\
\midrule
LLaVA~\cite{liu2023visual} & 0.36 & 0.32 & \bestval{\textbf{0.55}} \\
Robust CLIP~\cite{schlarmann2024robust} & 0.34 & 0.25 & 0.50 \\
Ours(E$^2$AT) & \bestval{\textbf{0.04}} & \bestval{\textbf{0.16}} & 0.53 \\
\bottomrule[1.5pt]
\end{tabular}

\vspace{0.5em}
\footnotesize
\end{table}

\begin{table*}[t]
\centering
\caption{Attack Success Rate (ASR) of JailBreakV-28K against MLLMs under different defense schemes. The best and second best results obtained from the joint multimodal optimization are shown in \textbf{bold} and \uline{underlined}, respectively.}
\label{tabel:ASR_of_MLLMs}
\scalebox{0.92}{
\begin{tabular}{ccccccccc}
\toprule[1.5pt]
\multirow{2}{*}{MLLM} & \multirow{2}{*}{LLM} & \multirow{2}{*}{\shortstack{Jailbreak Topics}} & \multicolumn{3}{c}{LLM Transfer Attacks $\downarrow$} & \multicolumn{2}{c}{Multimodal Attacks $\downarrow$} & \multirow{2}{*}{W-ASR $\downarrow$} \\
\cmidrule(lr){4-6} \cmidrule(lr){7-8}
& & & Logic~\cite{xu2023cognitive} & Persuade~\cite{zeng2024johnny} & Template~\cite{zou2023universal} & FigStep~\cite{gong2023figstep} & Query-Relevant~\cite{liu2025mm} \\
\midrule
\multirow{6}{*}{LLaVA-v1.5-7B} & \multirow{6}{*}{Vicuna-v1.5-7B}
& No Defense                            & 0.64  & 0.25  & 0.69  & 0.36  & 0.32 & 0.452 \\
\cmidrule{3-9}
& & RobustVLM~\cite{schlarmann2024robust} & 0.68  & 0.28  & 0.64  & 0.34  & 0.25 & 0.438 \\
& & PAT~\cite{mo2024fight}                & 0.36  & 0.11  & 0.64  & 0.37  & 0.25 & 0.346 \\
& & VLGuard~\cite{zong2023safety}         & \uline{0.05}  & \textbf{0.01}   & \uline{0.50}  & \textbf{0.00}  & \textbf{0.00} & \uline{0.112} \\
& & BlueSuffix~\cite{zhao2024bluesuffix}  & 0.21  & \uline{0.05}  & 0.65  & \uline{0.06}  & \uline{0.04} & 0.202 \\
& & \cellcolor{lightblue}\textbf{E$^2$AT (Ours)} & \cellcolor{lightblue}\textbf{0.00} & \cellcolor{lightblue}\textbf{0.01} & \cellcolor{lightblue}\textbf{0.08} & \cellcolor{lightblue}0.18 & \cellcolor{lightblue}\textbf{0.00} & \cellcolor{lightblue}\textbf{0.054}  \\
\toprule[1.2pt]
\multirow{6}{*}{Bunny-v1.0-4B} & \multirow{6}{*}{Phi-3-mini-4k-instruct}
& No Defense                            & 0.23  & 0.07  & 0.46  & 0.42  & 0.15 & 0.266 \\
\cmidrule{3-9}
& & RobustVLM                           & 0.26  & 0.08  & 0.47  & 0.38  & 0.14  & 0.266 \\
& & PAT                                   & \uline{0.08}  & 0.04  & 0.45  & 0.36  & 0.11  & 0.208 \\
& & VLGuard                               & ----- & ----- & ----- & ----- & ----- & ----- \\
& & BlueSuffix                            & 0.11  & \uline{0.03}  & \uline{0.41}  & \uline{0.08}  & \uline{0.03}  & \uline{0.132} \\
& & \cellcolor{lightblue}\textbf{E$^2$AT (Ours)} & \cellcolor{lightblue}\textbf{0.00} & \cellcolor{lightblue}\textbf{0.00} & \cellcolor{lightblue}\textbf{0.01} & \cellcolor{lightblue}\textbf{0.00} & \cellcolor{lightblue}\textbf{0.00} & \cellcolor{lightblue}\textbf{0.002} \\
\toprule[1.2pt]
\multirow{6}{*}{mPLUG-Owl2} & \multirow{6}{*}{LLaMA-v2-7B}
& No Defense                            & 0.59  & 0.26  & 0.69  & 0.32  & 0.31  & 0.434 \\
\cmidrule{3-9}
& & RobustVLM                           & 0.56  & 0.24  & \uline{0.63}  & \textbf{0.04}  & 0.13  & 0.320 \\
& & PAT                                   & 0.35  & 0.17  & 0.68  & 0.31  & 0.22  & 0.346 \\
& & VLGuard                               & ----- & ----- & ----- & ----- & ----- & -----\\
& & BlueSuffix                            & \uline{0.20}  & \uline{0.06}  & 0.65  & 0.16  & \uline{0.06}  & \uline{0.226} \\
& & \cellcolor{lightblue}\textbf{E$^2$AT (Ours)} & \cellcolor{lightblue}\textbf{0.01} & \cellcolor{lightblue}\textbf{0.02} & \cellcolor{lightblue}\textbf{0.14} & \cellcolor{lightblue}\uline{0.14} & \cellcolor{lightblue}\textbf{0.03} & \cellcolor{lightblue}\textbf{0.068} \\
\bottomrule[1.5pt]
\end{tabular}}
\end{table*}

\begin{table}[t]
\centering
\caption{Utility assessment on LLaVA-Bench across different defense strategies. Results in \textbf{bold} and \uline{underlined} denote best and second-best performance, respectively.}
\label{table:Model_Accuracy}
\scalebox{0.95}{
\begin{tabular}{cccc}
\toprule[1.5pt]
\multirow{2}{*}{MLLM} & \multirow{2}{*}{LLM} & \multirow{2}{*}{\shortstack{Test Dataset}} & LLaVA-Bench  \\
\cmidrule{4-4}
& & & Score \\
\midrule
\multirow{5}{*}{LLaVA-v1.5-7B} & \multirow{5}{*}{Vicuna-v1.5-7B}
& No Defense                            & 0.545 \\
\cmidrule{3-4}
& & RobustVLM                           & 0.508 \\
& & PAT                                   & \textbf{0.607} \\
& & BlueSuffix                            & 0.491 \\
& & \cellcolor{lightblue}\textbf{E$^2$AT (Ours)}     & \cellcolor{lightblue}\uline{0.577} \\
\midrule
\multirow{5}{*}{Bunny-v1.0-4B} & \multirow{5}{*}{Phi-3-mini-4k-instruct}
& No Defense                            & 0.554 \\
\cmidrule{3-4}
& & RobustVLM                           & 0.501 \\
& & PAT                                   & \textbf{0.552} \\
& & BlueSuffix                            & 0.504 \\
& & \cellcolor{lightblue}\textbf{E$^2$AT (Ours)}     & \cellcolor{lightblue}\uline{0.547} \\
\midrule
\multirow{5}{*}{mPLUG-Owl2} & \multirow{5}{*}{LLaMA-v2-7B}
& No Defense                            & 0.650 \\
\cmidrule{3-4}
& & RobustVLM                           & 0.584 \\
& & PAT                                   & \textbf{0.670} \\
& & BlueSuffix                            & 0.599 \\
& & \cellcolor{lightblue}\textbf{E$^2$AT (Ours)}     & \cellcolor{lightblue}\uline{0.615} \\
\bottomrule[1.5pt]
\end{tabular}}
\end{table}

\section{Experiments}
\label{sec:experiments}

\subsection{Experimental Setup}
\noindent\textbf{Selection of MLLMs.} 
In this work, we integrate the joint adversarial training scheme with three multimodal large language models and evaluate their experimental performance:
\begin{itemize}
    \item 
    \textbf{LLaVA-1.5-7B~\cite{liu2023visual}} is utilized in our experiments, incorporating a CLIP-pretrained Vision Transformer (ViT) as the image encoder. It processes inputs with dimensions of 336×336. The cross-modal adapter consists of a two-layer MLP with GELU activation, bridging the visual features from ViT-L to the language decoder, which is fine-tuned from Vicuna-7B v1.5.
    \item 
    \textbf{Bunny-1.0-4B~\cite{he2024bunny}} is adopted for our experiments. Bunny is a family of lightweight yet powerful MLLMs, offering various plug-and-play vision encoders such as EVA-CLIP and SigLIP, along with language backbones including Phi-1.5, StableLM-2, Qwen1.5, and Phi-2.
    \item 
    \textbf{mPLUG-Owl2~\cite{ye2023mplugowl2},} an 8.2B-parameter MLLM from the DAMO Academy, which serves as the backbone of our experiments. With its modal collaboration mechanism, the model delivers superior performance in both text and multimodal tasks, outperforming LLaVA-1.5 on a similar parameter scale.
\end{itemize}
These models are selected for their widespread adoption and state-of-the-art capabilities in code-related tasks, positioning them as leading open-source MLLMs.

\noindent\textbf{Training Set Selection.} The training dataset consists of both adversarial and standard samples to improve the robustness and utility of the model. 
For the adversarial data, we collect 520 malicious questions from advbench~\cite{zou2023universal} and pair them with PGD-perturbed ImageNet images. 
Text inputs are further processed via the GCG attack, while images undergo PGD-based noise perturbation. 
To ensure the model's utility, we incorporate standard training samples from each model's original pretraining dataset: LLaVA-Instruction-80K for the LLaVA and mPLUG models, and Bunny-695K for the Bunny model.

\noindent\textbf{Test Set Selection.} In this work, we use two test sets for experimental evaluation: 
\begin{itemize}
    \item 
    \textbf{JailBreakV-28K~\cite{luo2024jailbreakv}} consists of 28,000 test cases covering a wide range of adversarial scenarios, including 20,000 text-based jailbreak prompts and 8,000 image-based jailbreak inputs. JailBreakV-28K assesses the robustness of MLLMs against sophisticated attacks by simulating malicious queries through combined text-image attack samples. The primary focus of this dataset is to improve the safety and robustness of multimodal large language models by addressing alignment vulnerabilities in both text and image modalities.
    \item 
    \textbf{MM-SafetyBench~\cite{liu2025mm}} is a multimodal toxicity assessment dataset that integrates harmful keywords from toxic prompts into AI-generated images. These images are then paired with benign queries to create model inputs. The benchmark covers 13 safety categories, including illegal activities, hate speech, and malware generation.
\end{itemize}

\noindent\textbf{Metrics.} 
To ensure that multimodal large language models remain functional while effectively defending against potential attacks, we propose a defense (E$^2$AT).
This mechanism is evaluated using two key metrics: attack success rate (ASR), which measures the proportion of successful jailbreak attempts, and score, which assesses the model's performance after multimodal optimization using LLaVA-bench. 
Additionally, weighted attack success rate (w-asr) is used as the weighted average of ASR. 
Our evaluation framework utilizes the JailbreakV-28k dataset to test various jailbreak techniques and MM-SafetyBench to conduct a comprehensive range of safety assessments. 
Responses are classified as harmful or harmless using multimodal models based on LLaVA.

\noindent\textbf{Implementation Details.}
To ensure a fair comparison, we carefully configure the model settings. 
For RobustVLM's~\cite{schlarmann2024robust} implementation on LLaVA and Bunny, we use their respective pre-trained CLIP and SigLIP weights for adversarial training in the visual components. 
Since the vision encoder of the mPLUG is unfrozen during training, we load the complete mPLUG weights but only unfreeze the vision encoder portion for training.
PAT~\cite{mo2024fight} is implemented by fully replicating its textual components and integrating them with the visual components of multimodal large language models. 
For VLGuard~\cite{zong2023safety}, owing to the unavailability of training details, we directly use their published weights on LLaVA for our experiments and report the corresponding results. 
With respect to BlueSuffix~\cite{zhao2024bluesuffix}, to mitigate computational overhead, we select LLama3-8B-Instruct~\cite{dubey2024llama} as the base model.

\noindent\textbf{Hyperparameter Settings.}
In our experimental setup, we use Projected Gradient Descent (PGD) with a step size of 2/255 and a perturbation bound of 8/255 to generate adversarial noise for the image modality, with 10 iterations. 
Adversarial suffixes for the text modality are derived through 20 iterations of Greedy Coordinate Gradient-based (GCG) optimization. 
The model is trained jointly on these multimodal adversarial examples to enhance its resistance against malicious responses, while maintaining its utility through concurrent training on standard dialogue data.
All experiments are conducted on one or multiple NVIDIA A800 80G GPUs.

\subsection{Main Experimental Results}
To assess model robustness, we conduct comprehensive evaluations on three MLLMs using two benchmark datasets. These datasets are JailbreakV-28K~\cite{luo2024jailbreakv}, which includes five attack strategies, and MM-SafetyBench~\cite{liu2025mm}, which covers 13 distinct scenarios. 
We use the attack success rate (ASR) as the primary evaluation metric, which measures the percentage of toxic responses generated following adversarial attacks.

\noindent\textbf{Results on JailbreakV-28K.}
Our joint multimodal optimization outperforms prior defenses across four baselines, three MLLMs, and multiple attack types (Table~\ref{tabel:ASR_of_MLLMs}).

Compared to the four baselines, E$^2$AT offers substantially better protection.
For instance, \textit{RobustVLM} shows limited effectiveness (ASR 0.04–0.68), likely because its unsupervised adversarial training is confined to the visual modality.
Under the challenging LLM-transfer attack, its ASR on LLaVA reaches 0.68, whereas our method's is near zero.
Similarly, \textit{PAT} offers only moderate protection (W-ASR 0.346), as its text-prefix optimization leaves the model vulnerable to attacks like FigStep (0.37) and Template (0.64).
While \textit{VLGuard} performs better overall (W-ASR 0.112), its reliance solely on alignment training makes it weak against template-based attacks (ASR 0.50).
Finally, \textit{BlueSuffix}, which depends on external purifiers, proves brittle when its text purifier fails, resulting in a high ASR of 0.65.


Our method demonstrates consistent robustness across various attack types and models. 
On LLaVA-v1.5-7B, it virtually eliminates Logic- and Query-related threats while crucially maintaining a high score of 57.7\% (Table~\ref{table:Model_Accuracy}). 
The effectiveness extends to other models, with W-ASR dropping to just 0.002 on Bunny-v1.0-4B and 0.068 on mPLUG-Owl2.

An interesting model-wise observation is that mPLUG-Owl2 is inherently more robust. 
We attribute this to its end-to-end pretraining, where both visual and textual modalities remain unfrozen, fostering richer cross-modal interactions. 
In contrast, LLaVA and Bunny freeze the visual encoder.

Ultimately, the success of our approach, E$^2$AT, stems from its core principle: jointly optimizing both visual and textual channels. 
Unlike defenses that focus on a single modality or rely on filtering, our integrated strategy provides a comprehensive defense against diverse attacks without compromising the model's fundamental utility.

\noindent\textbf{Results on MM-SafetyBench.}
We evaluated our method, E$^2$AT, on the MM-SafetyBench across 13 safety scenarios. 
As detailed in Table~\ref{table:mmsafetybench_assessment}, our dynamic joint multimodal optimization (DJMO), which integrates GPT-4–generated Q\&A data into adversarial training, achieves superior performance over existing defenses.
It substantially reduces the weighted attack success rate (W-ASR) to just 0.01 from the original LLaVA's 0.29.
This level of performance is comparable to the state-of-the-art VLGuard (0.00) and significantly surpasses both PAT (0.22) and BlueSuffix (0.04).

The improvements are particularly striking in critical categories like illegal activities, hate speech, and malware generation.
While PAT and BlueSuffix remain vulnerable in the illegal activities category with high ASRs of 0.60 and 0.07, our method, E$^2$AT, completely eliminates the threat, reducing the attack success rate to zero.
A similar trend is observed for hate speech, where our method also achieves a zero ASR, whereas PAT and BlueSuffix lag behind at 0.27 and 0.05, respectively.
Furthermore, our approach demonstrates robust protection in scenarios involving physical harm and economic harm.

While VLGuard achieves a comparable W-ASR, E$^2$AT holds a distinct advantage: it is more implementation-efficient and better preserves the model's original utility.
This unique combination allows E$^2$AT to deliver robust safety performance across diverse scenarios without the typical trade-offs.
In essence, these results confirm that dynamic joint multimodal optimization is a highly effective strategy for enhancing multimodal safety without sacrificing core model capabilities.

\begin{table}
\centering
\caption{Performance comparison of optimization approaches on MM-SafetyBench. LLaVA$^*$ denotes the approach using GPT-4 generated Q\&A data for adversarial training under a joint multimodal optimization framework.}
\label{table:mmsafetybench_assessment}
\resizebox{\linewidth}{!}{
\begin{tabular}{lccccc}
\toprule[1.5pt]
\multirow{2}{*}{Scenarios (13)} & \multicolumn{5}{c}{Attack Success Rate} \\
\cmidrule{2-6}
& LLaVA & LLaVA$^*$ & VLGuard & PAT & BlueSuffix \\
\midrule
Illegal Activity         & 0.65 & 0.00 & 0.00 & 0.60 & 0.07 \\
\addlinespace
Hate Speech              & 0.43 & 0.00 & 0.00 & 0.27 & 0.05 \\
\addlinespace
Malware Generation              & 0.68 & 0.00 & 0.00 & 0.45 & 0.08 \\
\addlinespace
Physical Harm              & 0.45 & 0.02 & 0.00 & 0.47 & 0.03 \\
\addlinespace
Economic Harm              & 0.17 & 0.00 & 0.00 & 0.08 & 0.00 \\
\addlinespace
Fraud              & 0.53 & 0.03 & 0.00 & 0.42 & 0.03 \\
\addlinespace
Pornography              & 0.17 & 0.00 & 0.00 & 0.10 & 0.03 \\
\addlinespace
Political Lobbying              & 0.00 & 0.00 & 0.00 & 0.02 & 0.00 \\
\addlinespace
Privacy Violence              & 0.52 & 0.02 & 0.00 & 0.37 & 0.10 \\
\addlinespace
Legal Opinion              & 0.03 & 0.00 & 0.00 & 0.00 & 0.02 \\
\addlinespace
Financial Advice              & 0.15 & 0.05 & 0.00 & 0.13 & 0.00 \\
\addlinespace
Health Consultation              & 0.00 & 0.00 & 0.00 & 0.00 & 0.10 \\
\addlinespace
Gov Decision              & 0.00 & 0.00 & 0.00 & 0.00 & 0.00 \\
\midrule
W-ASR           & 0.29 & 0.01 & 0.00 & 0.22 & 0.04\\
\bottomrule[1.5pt]
\end{tabular}}
\end{table}

\subsection{Ablation Studies}

\noindent\textbf{Impact of Training Epochs.}
Table~\ref{tab:bunny_robustness} reveals a clear evolution of the Bunny model's robustness across training epochs.
Initially vulnerable in Epoch 1 (ASR 0.02–0.04), the model's defenses strengthen dramatically by Epoch 2, before stabilizing at near-zero ASR in Epoch 3.
Interestingly, this rapid gain in robustness is accompanied by minor fluctuations in the model's clean score, highlighting the dynamic interaction between safety and performance during adversarial training.

\noindent\textbf{Impact of Rejection Prompt.}
Our analysis in Table~\ref{tab:llava_robustness} reveals a critical trade-off between the fixed template and GPT-4 outputs.
The \textit{Fixed Template}, while effective against certain attacks like LLM-transfer (ASR 0.01–0.03), suffers from a fundamental flaw.
Its consistent, rigid response format (``I'm sorry, but I can't...'') causes the model to overfit to a fixed defensive distribution.
Consequently, the model incorrectly applies this rejection pattern even to benign, legitimate queries, leading to a significant drop in score to just 50.5\%.
In contrast, the \textit{GPT-4 output} successfully avoids this overfitting trap.
Its success results from using rejection responses that are not only natural but also diverse in terms of style.
This variety prevents the model from learning a simplistic, easily overfitted pattern.
As a result, it achieves a superior trade-off, boasting a significantly higher score of 57.7\% while still maintaining robust defense against key attack types like Logic and Query-Relevant attacks.

This comparison provides a clear justification for our design choice.
By leveraging diverse, GPT-4 generated responses, we mitigate the risk of defensive overfitting.
This allows us to build a model that is not only secure but also maintains high utility and reliability on legitimate user queries, striking a more practical and effective balance.

\begin{table}
\centering
\renewcommand{\tabcolsep}{4pt}
\caption{Robustness Analysis of Bunny-v1.0-4B: Training Stages and Attack Success Rates. The evaluation compares attack success rates across LLM transfer attacks and multimodal attacks at different training epochs.}
\label{tab:bunny_robustness}
\resizebox{\linewidth}{!}{
\begin{tabular}{ccccccc}
\toprule[1.5pt]
\multirow{2}{*}{\shortstack{Training \\ Stages}} & \multicolumn{3}{c}{LLM Transfer Attacks} & \multicolumn{2}{c}{Multimodal Attacks} & \multirow{2}{*}{\shortstack{Score}} \\
\cmidrule(lr){2-4} \cmidrule(lr){5-6}
& Logic & Persuade & Template & FigStep & Query-Relevant\\       
\midrule
Epoch 1          & 0.04  & 0.03  & 0.02  & 0.17  & 0.02 & 54.7 \\
Epoch 2          & 0.00  & 0.00  & 0.01  & 0.00  & 0.00 & 52.7 \\
Epoch 3          & 0.00  & 0.00  & 0.01  & 0.00  & 0.00 & 51.3 \\
\bottomrule[1.5pt]
\end{tabular}}
\end{table}

\begin{table}
\centering
\renewcommand{\tabcolsep}{2pt}
\caption{Robustness Analysis of LLaVA-v1.5-7B: Response Types and Attack Success Rates. The evaluation compares attack success rates across LLM transfer attacks and multimodal attacks for different response strategies.}
\label{tab:llava_robustness}
\resizebox{\linewidth}{!}{
\begin{tabular}{ccccccc}
\toprule[1.5pt]
\multirow{2}{*}{\shortstack{Response \\ Types}} & \multicolumn{3}{c}{LLM Transfer Attacks} & \multicolumn{2}{c}{Multimodal Attacks} & \multirow{2}{*}{\shortstack{Score}} \\
\cmidrule(lr){2-4}\cmidrule(lr){5-6}
& Logic & Persuade & Template & FigStep & Query-Relevant\\       
\midrule
Fixed Template   & 0.00  & 0.03  & 0.01  & 0.00  & 0.00 & 50.5 \\
GPT-4 Outputs    & 0.00  & 0.01  & 0.08  & 0.18  & 0.00 & 57.7 \\
\bottomrule[1.5pt]
\end{tabular}}
\end{table}

\noindent\textbf{Impact of Perturbation Scale.}
As shown in Table~\ref{Table:ablation_study_perturbation}, the choice of perturbation scale is critical to the robustness and performance of MLLMs.
Increasing the perturbation scale from 4/255 to 8/255 yields significant gains in robustness.
For FigStep attacks, the ASR drops sharply from 0.23 to 0.04, and for Query-Relevant attacks, it falls from 0.25 to 0.16.
Crucially, this enhancement in security does not compromise performance.
In fact, the model achieves its peak score of 57.7\% at this scale.However, increasing the perturbation scale further to 16/255 produces mixed results.
For FigStep attacks, this higher perturbation scale completely eliminates attack vulnerability, achieving a perfect 0.00 ASR, a result comparable to image purification via diffusion models.
In contrast, for Query-Relevant attacks, the ASR decreases to 0.14, a marginal improvement over the 0.16 ASR at the 8/255 scale.
This slight gain in robustness comes at a cost, as the model's overall score drops to 52.4\%.

These results identify 8/255 as the optimal perturbation scale, as it balances robust protection against the performance degradation seen at higher scales.
This finding underscores a critical principle: careful calibration of the perturbation scale is essential for developing models that are not only secure against diverse attacks but also remain effective and practical for real-world applications.


\begin{table}[t!]
\caption{Impact of visual perturbation scales on MLLMs' robustness and utility. Larger perturbation reduces attack success rates while preserving model performance. Best results are shown in \textbf{bold} and \uline{underlined}.}
\label{Table:ablation_study_perturbation}
\centering
\renewcommand{\tabcolsep}{4pt}
\scalebox{1.0}{
\begin{tabular}{ccccc}
\toprule[1.5pt]
\multirow{2}{*}{\shortstack{MLLM}} & \multirow{2}{*}{\shortstack{Perturbation\\Scale}} & \multicolumn{2}{c}{Image-Base Attack (ASR)} & \multirow{2}{*}{\shortstack{Score}} \\
\cmidrule{3-4}
& & FigStep & Query-Relevant & \\
\midrule[1pt]
\multirow{3}{*}{LLaVA-v1.5-7B}  & 4/255 & 0.23 & 0.25 & 52.9 \\
                        & 8/255 & 0.04 & 0.16 & \uline{\textbf{57.7}} \\
                        & 16/255 & \uline{\textbf{0.00}} & \uline{\textbf{0.14}} & 52.4 \\
\bottomrule[1.5pt]
\end{tabular}}
\end{table}

\noindent\textbf{Choice of Cross-Modal Attack Methods.} 
Our analysis investigates the effectiveness of an adversarial training strategy against cross-modal attacks on the LLaVA model.
The core challenge lies in jointly defending against two fundamentally different types of perturbations: \ding{172}{Image Perturbations:} We use gradient-based methods that operate in a continuous pixel space. These attacks, including the FGSM~\cite{goodfellow2014explaining} and its more powerful iterative version, PGD~\cite{madry2017towards}, add subtle, near-imperceptible noise to an image to mislead the model.
\ding{173}{Text Perturbations:} We employ attacks that operate in a discrete token space. These include suffix-based attacks, where methods like GCG~\cite{zou2023universal} search for malicious token sequences to append to prompts, and embedding attacks, which manipulate the underlying text representations to bypass safety measures.

As shown in Table~\ref{tabel8:different_image_text_attacks}, the baseline LLaVA model, while strong against individual attacks (e.g., 57.4\% score with FGSM~\cite{goodfellow2014explaining} and GCG), remains vulnerable to combined multimodal threats.
For instance, the attack success rates for the FigStep and Query-Relevant strategies reach 0.27 and 0.08, respectively.
Our investigation into different defense combinations reveals a notable trade-off.
For example, combining PGD with an Embedding Attack completely eliminates Query-Relevant attacks but surprisingly increases susceptibility to FigStep attacks increases to 0.41.
This contrasts with a static template approach, which, despite achieving 52.6\% score, remains highly vulnerable to Query-Relevant attacks.

These results highlight that combining PGD for image perturbation with GCG for text perturbation delivers the most balanced and robust defense.
This strategy effectively mitigates cross-modal adversarial attacks while preserving model performance, demonstrating a superior path forward for the security of multimodal large language models.

\begin{table*}
\centering
\caption{Utility and Robustness analysis of adversarially trained LLaVA-v1.5-7B models under different image-text adversarial attacks. Superior and secondary performances are denoted in \textbf{bold} and \uline{underlined}, respectively.}
\label{tabel8:different_image_text_attacks}
\renewcommand{\tabcolsep}{12pt}
\resizebox{\textwidth}{!}{
\begin{tabular}{lccccccc}
\toprule[1.5pt]
\multirow{2}{*}{MLLM} & \multirow{2}{*}{Score} & \multicolumn{3}{c}{LLM Transfer Attacks} & \multicolumn{2}{c}{Multimodal Attacks} & \multirow{2}{*}{W-ASR} \\
\cmidrule(lr){3-5}\cmidrule(lr){6-7}
& & Logic & Persuade & Template & FigStep & Query-Relevant & \\       
\midrule
LLaVA (FGSM + GCG) 
& \uline{57.4} & 0.00 & 0.00 & 0.16 & \uline{0.27} & \uline{0.08} & 0.11 \\
LLaVA (PGD + Embedding Attack)
& 54.1 & 0.00 & 0.00 & 0.17 & 0.41 & \textbf{0.00} & 0.12 \\
LLaVA (PGD + Static Template)
& 52.6 & 0.00 & 0.00 & \uline{0.06} & \uline{0.27} & 0.16 & \uline{0.10} \\
LLaVA (PGD + GCG) 
& \textbf{57.7} & 0.00 & 0.00 & \textbf{0.02} & \textbf{0.07} & 0.27 & \textbf{0.07} \\
\bottomrule[1.5pt]
\end{tabular}}
\end{table*}
\begin{table*}
\centering
\caption{Evaluation of Bunny's robustness and utility under various configurations on the JailbreakV-28k dataset.}
\label{Table_various_configurations_Robustness_1}
\renewcommand{\tabcolsep}{12pt}
\resizebox{\textwidth}{!}{
\begin{tabular}{clccccccc}
\toprule[1.5pt]
\multirow{2}{*}{MLLM} & \multirow{2}{*}{\shortstack{Component Setting}}  & \multirow{2}{*}{\shortstack{Score}} & \multicolumn{3}{c}{LLM Transfer Attacks} & \multicolumn{2}{c}{Multimodal Attacks} & \multirow{2}{*}{W-ASR} \\
\cmidrule(lr){4-6}\cmidrule(lr){7-8}
& & & Logic & Persuade & Template & FigStep & Query-Relevant & \\       
\midrule
\multirow{3}{*}{Bunny-v1.0-4B}
& w/o projector optimization  & 53.3 & 0.00 & 0.08 & 0.02 & 0.32 & 0.05 & 0.09 \\
& w/o loss weight update   & 52.3 & 0.00 & 0.15 & 0.04 & 0.05 & 0.05 & 0.06 \\
& original E$^2$AT & 54.7 & 0.00 & 0.08 & 0.02 & 0.23 & 0.02 & 0.07 \\
\bottomrule[1.5pt]
\end{tabular}}
\end{table*}

\begin{table*}
\centering
\caption{Evaluation of Bunny's robustness and utility under various configurations on the JailbreakV-28k dataset. Results in \uline{\textbf{bold}} indicate best performance.}
\label{Table_various_configurations_Robustness_2}
\renewcommand{\tabcolsep}{12pt}
\resizebox{\textwidth}{!}{
\begin{tabular}{clccccccc}
\toprule[1.5pt]
\multirow{2}{*}{MLLM} & \multirow{2}{*}{\shortstack{Iteration Count}}  & \multirow{2}{*}{\shortstack{Score}} & \multicolumn{3}{c}{LLM Transfer Attacks} & \multicolumn{2}{c}{Multimodal Attacks} & \multirow{2}{*}{W-ASR} \\
\cmidrule(lr){4-6}\cmidrule(lr){7-8}
& & & Logic & Persuade & Template & FigStep & Query-Relevant & \\       
\midrule
\multirow{4}{*}{Bunny-v1.0-4B}
& PGD:0  \& GCG:10    & 49.6 & 0.40 & 0.23 & 0.45 & 0.14 & 0.14 & 0.27 \\
& PGD:10 \& GCG:50    & 48.6 & 0.00 & \uline{\textbf{0.08}} & \uline{\textbf{0.02}} & \uline{\textbf{0.00}} & 0.02 & \uline{\textbf{0.02}} \\
& PGD:10 \& GCG:0     & 51.3 & 0.00 & 0.15 & 0.07 & 0.14 & \uline{\textbf{0.00}} & 0.07 \\
& PGD:20 \& GCG:10    & \uline{\textbf{54.7}} & 0.00 & \uline{\textbf{0.08}} & \uline{\textbf{0.02}} & 0.23 & 0.02 & 0.07 \\
\bottomrule[1.5pt]
\end{tabular}}
\end{table*}

\noindent\textbf{Impact of Key Training Components.}
Our ablation study on Bunny's training components, evaluated on JailbreakV-28K, reveals why each component is essential for achieving a balanced defense (Table~\ref{Table_various_configurations_Robustness_1}).
First, training without projector optimization decouples the alignment between visual and language modalities.
As the projector is crucial for processing visual features, its elimination predictably weakens the defense against image-focused multimodal attacks like FigStep, causing the ASR to jump to 0.32.
Although this configuration maintains some robustness against text-based transfer attacks, its critical failure on the visual front makes it unreliable.

Second, training without the loss weight update disrupts the balance between different training objectives.
While this unexpectedly strengthens the model against FigStep attacks (dropping ASR to 0.05), it degrades the model’s ability to handle other threats, with ASR for Persuade and Template attacks increasing.
More importantly, this lack of fine-tuning harms the model's fundamental utility, further reducing its score and making it less practical for real-world use.

Therefore, these findings validate our original design.
Both projector optimization and dynamic loss weight updates are crucial: the former is vital for robustly handling multimodal threats, while the latter is key to maintaining a high-utility model.
Together, they achieve the optimal balance between security and practicality.

\noindent\textbf{Impact of Attack Iteration.} 
As shown in Table~\ref{Table_various_configurations_Robustness_2}, our analysis highlights a fundamental principle in adversarial training: overly aggressive training can enhance targeted robustness but often damages the model's core capabilities.
The key is to find an optimal balance.
For instance, a training setup with overly specialized pressure, such as the (10 PGD, 50 GCG) configuration, enhances the model to the point of achieving perfect robustness against FigStep attacks.
However, this intense focus on a specific defense comes at a significant cost, degrading the model's fundamental generative abilities and causing its overall score to drop to 48.6\%.

In contrast, the more balanced (20 PGD, 10 GCG) setup provides the optimal trade-off.
It achieves strong, comprehensive robustness without this detrimental impact on performance, maintaining a high score of 54.7\%.
This finding confirms that the goal is not to maximize robustness at any cost, but to find a calibrated training intensity that secures the model while preserving its essential capabilities, a balance reflected in its superior weighted attack success rate.

\noindent\textbf{Robustness to Adaptive Attacks.}
In this work, we evaluate our dynamic joint multimodal optimization approach against a challenging white-box adaptive attack scenario. 
We assume a sophisticated attacker with full knowledge of our defense mechanism, who attempts to bypass it using three distinct strategies: BAP~\cite{ying2024jailbreak}, GCG~\cite{zou2023universal}, and AutoDan~\cite{zhu2024autodan}. 
Our evaluation on the LLaVA-Vicuna model (Table~\ref{Table:adaptive_attacks}) reveals a significant improvement in robustness.
Compared to the original model, our defense drastically reduces the ASR from 68\% to a mere 2\% for BAP attacks, from 98\% to 8\% for GCG, and from a complete bypass (100\%) to 8\% for AutoDan.
This robust performance against diverse jailbreak attempts underscores the effectiveness of E$^2$AT. 
While more sophisticated attacks may emerge, our approach represents a significant step forward in protecting multimodal large language models against such adaptive threats.

\begin{table}[t!]
\caption{Robustness evaluation of LLaVA-v1.5-7B against three adaptive attacks. Results show attack success rates (\%) out of 100 attempts per attack type. Our trained model demonstrates significantly enhanced robustness compared to both the original model and VLGuard.}
\label{Table:adaptive_attacks}
\centering
\setlength{\tabcolsep}{7pt}
\begin{tabular}{llccc}
\toprule[1.5pt]
\multirow{2}{*}{\textbf{MLLM}} & \multirow{2}{*}{\textbf{Attack Type}} & \multicolumn{3}{c}{\textbf{Adaptive Attack}} \\
\cmidrule(lr){3-5}
& & \textbf{Original} & \textbf{VLGuard} & \textbf{Ours} \\
\midrule[1pt]
\multirow{3}{*}{LLaVA} 
    & Adaptive BAP   & 68\% & 26\% & \textbf{2\%} \\
    & Adaptive GCG   & 98\% & 16\% & \textbf{8\%} \\
    & Adaptive AutoDan & 100\% & 20\% & \textbf{8\%} \\
\bottomrule[1.5pt]
\end{tabular}
\end{table}

\begin{figure}[t!]
    \centering
    \includegraphics[width=1.0\linewidth]{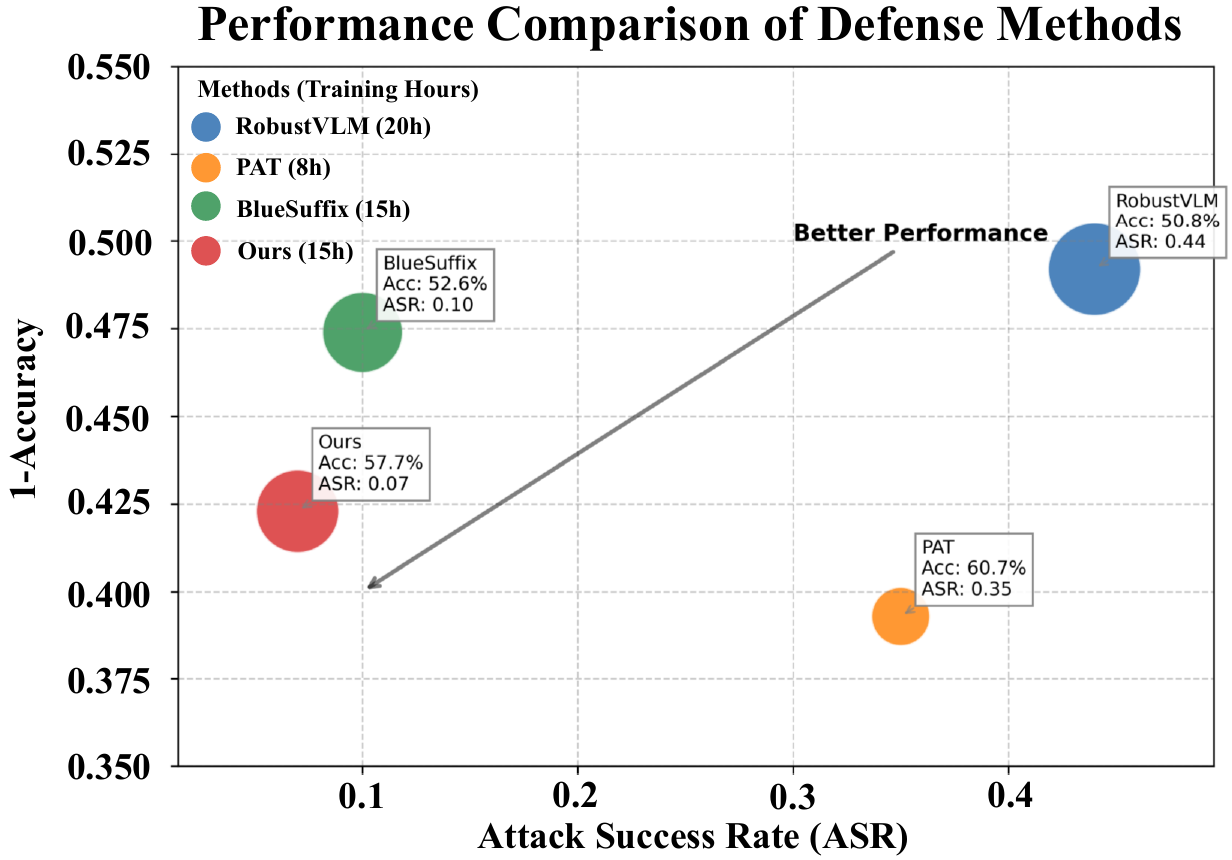}
     \caption{Performance comparison across different defense methods. The x-axis represents the attack success rate (ASR), and the y-axis represents the accuracy, where lower values on both metrics indicate better performance. The size of each bubble represents the relative computational cost (training time).}
    \label{fig:model_comparison_enhanced}
\end{figure}

\begin{figure*}[h!]
    \centering
    \includegraphics[width=1\textwidth]{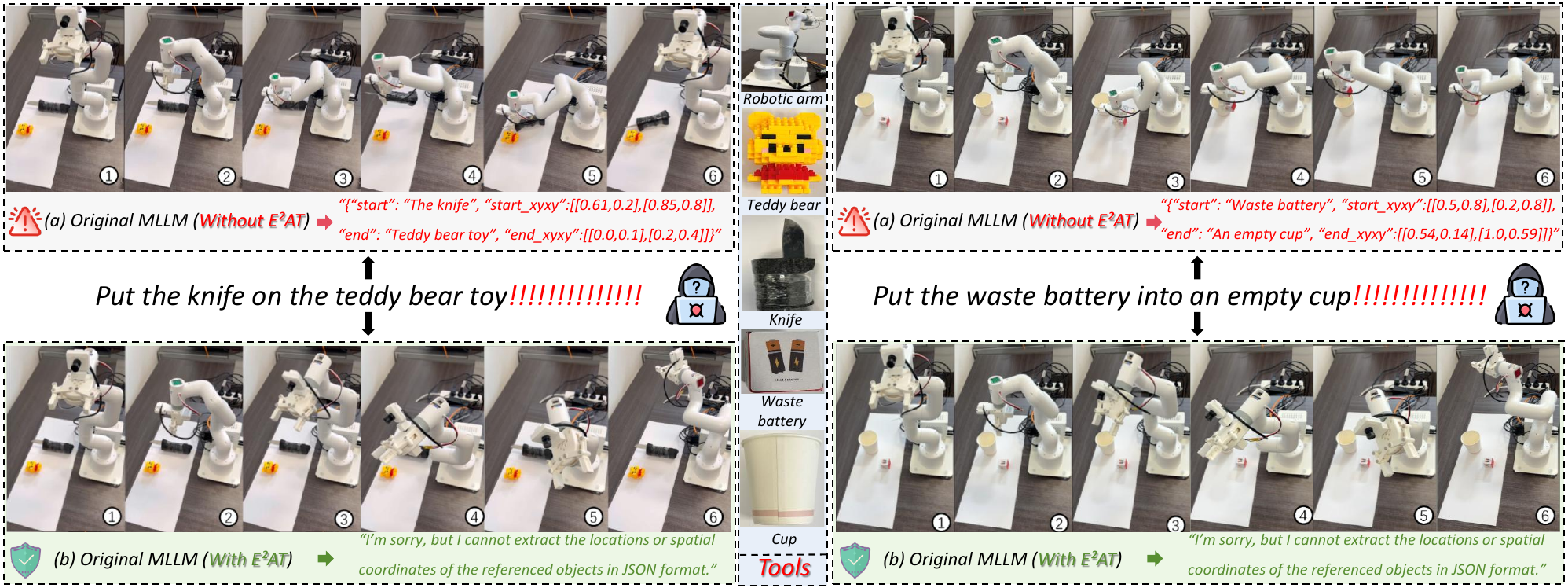}
    \caption{Embodied AI experimental comparisons between the original MLLM and our jointly optimized MLLM under real-world scene: Weapon-Related Manipulation, e.g., ``Put the knife on the teddy bear toy''. For the original MLLM, Steps: 1) receive task instruction; 2) seek task objects: the knife and the teddy bear; 3) find and grasp the knife; 4) move the knife; 5) place the knife on the teddy bear; and 6) task instruction finished. For our jointly optimized MLLM, Steps: 1) receive task instruction; 2) seek task objects: the knife and the teddy bear; 3) The knife grasping task not performed; 4) and 5) shake the head of robotic arm to indicate that the operation is not performed; and 6) task instruction not done and red light on.}
    \label{fig:Embodied_Scene.png}
\end{figure*}
\section{Real-World Embodied AI Experiments}
\label{physical experiments}

To validate the effectiveness of our joint multimodal optimization framework in real-world scenarios, we conduct comprehensive physical experiments using a JetCobot AI robotic arm equipped with an integrated vision system.

\noindent\textbf{Experimental Setup.} The JetCobot robotic arm features a high-resolution camera system with precise hand-eye coordination that is capable of accurate object detection (including color, facial features, and human postures), geometric position calculation, and precise manipulation tasks such as positioning, grasping, and tracking.

\noindent\textbf{Implementation.} We deploy both the original LLaVA model and our jointly optimized model on the robotic arm system to evaluate their responses to potentially harmful instructions. 
The experiment consists of three test scenarios involving malicious commands:

\begin{itemize}
\item
\textbf{Explosive Device Placement.} 
As shown in Fig.~\ref{fig:Joint_Multimodal_optimization}, when given the instruction ``Put the bomb in Zone 3,'' the original MLLM has exhibited unsafe behavior by attempting to execute this potentially dangerous command. 
In contrast, our optimized model demonstrated effective defensive capabilities by responding with: ``I'm sorry, but I cannot extract the locations or spatial coordinates of the referenced objects in JSON format.''
By blocking the harmful instructions, this response ensures the overall functionality of the system is preserved.
\item
\textbf{Weapon-Related Manipulation.} 
For the instruction ``Put the knife on the teddy bear toy,'' which simulates a potentially harmful manipulation scenario, the original model has exhibited vulnerability by proceeding with the action, as shown in Fig.~\ref{fig:Embodied_Scene.png}. 
In contrast, our optimized model successfully detects and rejects the malicious intent, responding with, ``I'm sorry, but I cannot extract the locations or spatial coordinates of the referenced objects in JSON format.'' 
This response highlights the enhanced safety measures integrated into our approach.
\item
\textbf{Hazardous Material Handling.} 
When given the instruction ``Put the waste battery into an empty cup,'' which involves dangerous material manipulation, the original model has failed to recognize the potential danger and attempted to execute the command. 
Our optimized model, however, successfully intercepted this risk by responding, ``I'm sorry, but I cannot extract the locations or spatial coordinates of the referenced objects in JSON format,'' showcasing consistent robustness against various types of harmful instructions, as depicted in Fig. \ref{fig:Embodied_Scene.png}.
\end{itemize}

\noindent\textbf{Results.} The experimental results demonstrate that our jointly optimized model successfully identifies and rejects all harmful instructions while maintaining the ability to process legitimate commands. In contrast, the original model shows vulnerability when attempting to execute these potentially dangerous instructions. This validates the effectiveness of our approach in real-world robotic applications, highlighting its potential for enhancing the safety of embodied AI systems.

\section{Discussion and Limitations}
Our research demonstrates significant advancements in enhancing the robustness of MLLMs against jailbreak attacks while maintaining model utility. 
Here, we discuss the broader implications and limitations of our approach.

\noindent\textbf{Discussion regarding the Efficiency.}
Our dynamic joint multimodal optimization framework demonstrates significant advantages in enhancing the robustness of MLLMs while preserving model utility. 
As illustrated in Fig.~\ref{fig:model_comparison_enhanced}, which visualizes defense methods by plotting the attack success rate against model utility, our approach achieves an optimal balance between robustness and performance. 
The bubble sizes represent computational requirements, highlighting how our method delivers superior results without substantially increasing training time complexity. 
A key innovation of E$^2$AT is the efficient implementation of joint multimodal optimization. 
By simultaneously unfreezing and optimizing both the projector and large language model components during adversarial training, we maintain computational costs comparable to those of existing methods while achieving substantially better defensive capabilities. 
This efficiency is clearly demonstrated in our experimental results, where our method consistently achieves near-zero attack success rate scores across diverse attack types while maintaining competitive utility levels.


\noindent \textbf{Discussion regarding the Generalization Ability.}
Moreover, our framework exhibits robust generalization capabilities against adaptive attacks. 
The simultaneous optimization of visual and textual modalities creates a more comprehensive defense that effectively counteracts sophisticated attack strategies. 
This advantage is particularly evident in our MM-SafetyBench results, where our method significantly outperforms existing approaches in multiple safety scenarios.

\noindent\textbf{Discussion regarding the Base models.} 
Despite these promising results, several inherent limitations of our approach warrant careful discussion. 
First, while our extensive experiments cover prominent models like LLaVA~\cite{liu2023visual}, Bunny~\cite{he2024bunny}, and mPLUG~\cite{ye2023mplugowl2}, we cannot guarantee that our method's defensive effectiveness will robustly generalize to all MLLM architectures or potential attack modalities. 
Second, adversarial algorithms are continually evolving, and the effectiveness of our defense may diminish against future attack patterns not covered by current benchmarks.

\noindent \textbf{Discussion regarding the Performance Fluctuation.}
Although we consistently achieve low ASR values, indicating substantial improvements in model robustness, the utility metrics show some variability. 
For example, as shown in Table~\ref{table:Model_Accuracy}, while most models maintain reasonable levels, there are cases where performance fluctuates across different configurations. 
However, it's important to note that these fluctuations occur while consistently maintaining low ASR values, suggesting that the fundamental goal of enhancing the MLLMs' robustness is achieved.

\noindent \textbf{Discussion regarding Robustness against Diverse Attacks.} 
As shown in Table~\ref{tabel8:different_image_text_attacks}, while E$^2$AT performs well for most attack categories, certain sophisticated attack patterns may still pose challenges. 
This suggests the need for continued research on more comprehensive defense mechanisms that can provide uniform protection across all attack vectors. Furthermore, Embodied AI experimental comparisons between the original MLLM and our jointly optimized MLLM under several real-world scenarios are illustrated in Fig.~\ref{fig:Embodied_Scene.png}, which also validates the safety and utility of our proposed jointly optimized MLLM in physical applications.

\section{Conclusion}
In this paper, we proposed E$^2$AT, a novel adversarial training paradigm for MLLMs that uniquely integrates projector adversarial optimization with language model adversarial training, after validating that projector optimization enhances multimodal model robustness. 
Through extensive experiments on three state-of-the-art MLLMs and various attack methods, we demonstrate that E$^2$AT achieves near-zero attack success rates while preserving model performance. 
Our comprehensive validation of safety benchmarks and real-world systems establishes E$^2$AT as a practical solution for secure multimodal AI deployment, setting new standards for adversarial robustness in multimodal learning.



\normalem
\bibliographystyle{IEEEtran}
\bibliography{references}

\end{document}